\documentclass[10pt,twocolumn,letterpaper]{article}

\usepackage{iccv}
\usepackage{times}
\usepackage{epsfig}
\usepackage{graphicx}
\usepackage{amsmath}
\usepackage{amssymb}

\usepackage[tableposition=above]{caption}
\usepackage{subcaption}
\usepackage{bbm}
\usepackage{bm}
\usepackage{xspace}
\usepackage[utf8]{inputenc} %
\usepackage{url}            %
\usepackage{booktabs}       %
\usepackage{amsfonts}       %
\usepackage{nicefrac}       %
\usepackage{microtype}      %
\usepackage{color}
\usepackage{algorithm}
\usepackage{algorithmic}
\usepackage{array}
\usepackage{multirow}
\usepackage[para,online,flushleft]{threeparttable}
\usepackage{float}
\usepackage{enumitem}
\usepackage{makecell}
\usepackage{gensymb}         %
\usepackage{mathtools}
\usepackage{booktabs}

\makeatletter
\DeclareRobustCommand\onedot{\futurelet\@let@token\@onedot}
\def\@onedot{\ifx\@let@token.\else.\null\fi\xspace}

\def\ie{\emph{i.e}\onedot}

\def\etal{\emph{et al}\onedot}
\makeatother

\usepackage{xcolor}
\usepackage{ifthen}
\usepackage{textcomp}
\definecolor{red}{rgb}{0.9,0.1,0}
\definecolor{slateblue}{rgb}{0.7,0.35,0.9}
\definecolor{green}{rgb}{0, 0.4, 0}
\definecolor{brown}{rgb}{0.3, 0.2, 0}
\definecolor{mahogany}{rgb}{0.75, 0.25, 0.0}
\definecolor{purple}{rgb}{0.3, 0, 0.3}
\definecolor{darkgreen}{rgb}{0, 0.4, 0}
\definecolor{frenchblue}{rgb}{0.0, 0.45, 0.73}
\definecolor{blue}{rgb}{0.0, 0.0, 1.0}
\definecolor{goldenrod}{rgb}{0.65, 0.45, 0.03}
\definecolor{gray}{rgb}{0.5,0.5,0.5}

\newcommand{\heading}[1]{\vspace{1mm}\noindent\textbf{#1}}
\newcommand{\red}[1]{{\color{red}#1}}
\newcommand{\blue}[1]{{\color{blue}#1}}
\newcommand{\best}[1]{{\red{\pmb{#1}}}}
\newcommand{\second}[1]{\blue{\underline{#1}}}

\newcommand{\tabref}[1]{Table~\ref{tab:#1}} %

 \definecolor{doraemonblue}{rgb}{0.20, 0.68, 0.77}

\definecolor{goldenrod}{rgb}{0.65, 0.45, 0.03}

\definecolor{MTKGold}{RGB}{241, 154, 33}

\newboolean{revising}
\setboolean{revising}{false}
\ifthenelse{\boolean{revising}}
{
    \newcommand{\ignore}[1]{}
    
    \newcommand{\yulunliu}[1]{{\color{red}#1}\normalfont}
    \newcommand{\albert}[1]{{\color{blue}#1}\normalfont}

    \newcommand{\ren}[1]{{\color{MTKGold}#1}\normalfont}

} {
    \newcommand{\ignore}[1]{}
    
    \newcommand{\yulunliu}[1]{#1}
    \newcommand{\albert}[1]{#1}

    \newcommand{\ren}[1]{#1}
}

\makeatletter
\renewcommand{\paragraph}{%
  \@startsection{paragraph}{4}%
  {\z@}{0.5\baselineskip \@plus 0ex \@minus 0ex}{-1em}%
  {\normalfont\normalsize\bfseries}%
}
\makeatother

\def\ie{\emph{i.e}\onedot}

\def\etal{\emph{et al}\onedot}
\makeatother

\usepackage[pagebackref=true,breaklinks=true,letterpaper=true,colorlinks,bookmarks=false]{hyperref}

\iccvfinalcopy %

\def\iccvPaperID{1939} %

\ificcvfinal\fi

\makeatletter
\let\@fnsymbol\@arabic
\makeatother

\begin{document}

\title{Bridging Unsupervised and Supervised Depth from Focus \\ via All-in-Focus Supervision}

\author{
Ning-Hsu Wang$^{1,2}$\quad\quad\quad
Ren Wang$^{1}$\quad\quad\quad
Yu-Lun Liu$^{1}$\quad\quad\quad
Yu-Hao Huang$^{1}$\\
Yu-Lin Chang$^{1}$\quad\quad\quad
Chia-Ping Chen$^{1}$\quad\quad\quad
Kevin Jou$^{1}$\\
\small$^{1}$MediaTek Inc. \quad\quad\quad
\small$^{2}$National Tsing Hua University\\
{\small\url{https://github.com/albert100121/AiFDepthNet}}
}

\maketitle
\ificcvfinal\thispagestyle{empty}\fi
\begin{abstract}
Depth estimation is a long-lasting yet important task in computer vision.
Most of the previous works try to estimate depth from input images and assume images are all-in-focus (AiF), which is less common in real-world applications.
On the other hand, a few works take defocus blur into account and consider \ren{it} as another cue for depth estimation.
In this paper, we propose a method to estimate \ren{not only} a depth map \ren{but} an AiF image from a set of images with different focus positions (known as a focal stack).
We design a shared \ren{architecture to exploit the relationship between depth and AiF estimation}.
\ren{As a result}, the proposed method can be trained \ren{either} supervisedly with ground truth depth, or \emph{unsupervisedly} \albert{with AiF} \ren{images as supervisory signals}.
We show in various experiments that our method outperforms the state-of-the-art methods both quantitatively and qualitatively\ren{, and also has higher efficiency in inference time}.
\end{abstract}

\noindent

\section{Introduction}

Depth estimation has been one of the most fundamental computer vision problems in decades.
Many downstream tasks, such as augmented reality (AR), virtual reality (VR) and autonomous driving, highly rely on this research topic.
Recently, it also \ren{enabled} an increasing number of applications for smartphone photography, such as depth-of-field adjustment, background substitution, and changing focus after the picture is taken.

Consequently, depth sensing has become a fundamental component for capturing devices. 
Active depth sensing solutions such as Time-of-Flight (ToF) and \albert{ structured light} are often expensive and power-consuming due to the need for specialized hardware. 
Passive techniques, such as binocular or multi-view stereo, are more cost and power-efficient but prone to errors in textureless regions. 

Deep learning based stereo matching \ren{methods} tackle this problem in a data-driven way by learning depth estimation directly from input images.
However, they require a large amount of high-quality paired training data, which are time-consuming and expensive to acquire.
They also suffer when the training data are imperfect: synthesized and unrealistic input images, or inaccurately registered depth maps.

Some unsupervised learning approaches \cite{MonoDepth, Deep3D} \ren{were} proposed to address this problem. They usually use image reconstruction loss and consistency loss without the need for ground truth depth data.
They can also mitigate domain gaps by training directly with real-world stereo images without corresponding registered depth maps.

Another relatively under-explored cue for depth estimation is defocus blur.
The task of depth-from-focus (or defocus) aims to estimate the depth of a scene from a focal stack, \ie, images taken at different focal positions by the same camera.
This allows consumer auto-focus monocular cameras to estimate depth without additional hardware.

Conventional optimization based depth-from-focus  approaches~\cite{MobilePhone,DBLP:conf/cvpr/SurhJPIHK17, VDFF} estimate the level of sharpness for each pixel and often suffer from \ren{textureless} objects or aperture problems. 
Deep learning techniques~\cite{DDFF, maximov2020focus} help to overcome these issues but need ground truth depth data for supervised \ren{learning}.  
It's difficult and time-consuming to retrieve focal stacks with registered depth maps, let alone the imperfect depth data obtained by hardware solutions such as ToF sensors~\cite{DDFF}. 
One could synthesize defocus blur on a synthetic dataset with synthetic depth \ren{maps}~\cite{maximov2020focus}. 
\albert{However, it is still questionable whether the thin lens synthesis model could represent real-world optics well.}

In this paper, we propose a novel method to estimate depth and an all-in-focus (AiF) image jointly from an input focal stack.
We exploit the relationship between these \ren{two tasks} %
and design a shared common network.
Moreover, the proposed network can be trained either supervisedly with ground truth depth maps or unsupervisedly with only ground truth AiF images.
Compared to high-quality \ren{labeled} depth, acquiring \ren{AiF images} is relatively \ren{easier} because AiF images can be captured with smaller apertures \ren{along with} longer exposures. \ren{However, collecting the corresponding focal stack of an AiF image might be difficult because of the focus breathing phenomenon, where a camera's field of view changes as the lens moves. To address this problem, we use synthetic data without such effects during training, and apply a calibration process on real data during testing.} %

Our \textbf{contribution} is three-fold:
\begin{itemize}
  \item When trained supervisedly, our method outperforms the state-of-the-art methods in various comparisons, while our method also runs faster.
  \item To our knowledge, the proposed method is the first that can learn depth estimation \ren{unsupervisedly from only AiF images} and performs favorably against the state-of-the-art methods. 
  \item Domain gaps can be mitigated by our method with \ren{test-time optimization} on real-world data, \ren{especially} when ground truth depth data are not available.
\end{itemize}

\section{Related Work}

\subsection{Depth from Focus}

Depth estimation is a fundamental \ren{computer vision task \albert{that} aims to use} %
different cues such as color, semantics, stereo, or \ren{the} difference in image sequences to predict \ren{or} fuse depth maps \cite{Dora3D, WangSTCS20}. 
Most of the previous works \ren{assume} that input images are all-in-focus, whereas in real-world scenarios,
\ren{images are usually} considered to be defocused in the background or with \ren{a shallow depth-of-field (DoF)}.
\ren{Nevertheless, some} approaches elaborate on depth estimation \ren{with} defocused images. Conventional optimization based approaches  \cite{MobilePhone,DBLP:conf/cvpr/SurhJPIHK17} \ren{proposed to directly} estimate depth from focal \ren{stacks, and a variant approach \cite{VDFF} tries} to generate an index map in which every pixel is assigned to \ren{the} focus position \ren{leading to the maximal sharpness}.
\ren{Chen \etal \cite{Blur-aware} found} the relationship between relative blur and disparity\ren{, and make use of it} to enhance the robustness \ren{of} matching.
Depth from focus sweep video \cite{DBLP:conf/3dim/KimRT16} targets estimating depth from images with successive focus positions.
\ren{Recently, deep} learning based approaches \cite{DDFF, DefocusNet} could model the blurriness more precisely and achieve much better depth quality.
\ren{On the other hand, some works \cite{DBLP:conf/bmvc/AnwarHP17a, DBLP:conf/icip/CarvalhoSTAC18, DBLP:conf/cvpr/SrinivasanGWNB18} use deep learning techniques to} remove the defocus blur for single \ren{images}.

\subsection{Multi-focus Image Fusion}

\ren{Although real-world images usually have defocus blur, most computer vision applications are supposed to cooperate with all-in-focus (AiF) images.} 
\ren{Therefore, more and more attention is given on shallow-DoF image deblurring and multi-focus image fusion}.
\ren{A multi-focus} image fusion approach~\cite{DBLP:journals/jcp/WangC11} \ren{applies} Laplacian to different scales of images.
Zhan \albert{\etal}~\cite{DBLP:journals/jihmsp/ZhanTLS15} \ren{proposed} a guided filter to help edge preserving during multi-focus image fusion.
Nejati \albert{\etal}~\cite{Nejati2015MultifocusIF} \ren{learn} a sparse representation of relative sharpness \ren{measurement and produce a pixel-level score map} for decision through pooling.
An unsupervised learning approach ~\cite{deepfuse} \ren{was} proposed to fuse either multi-exposure or multi-focus images to generate a \ren{high dynamic range (HDR) or an AiF image}. Liu \albert{\etal}~\cite{liu2017multi} \ren{proposed to learn a focus map and a segmentation map through deep neural networks, and fuse images by integrating these maps}.

\subsection{Light Field}

A light field camera captures spatially distributed light rays. By re-rendering \ren{through} digitized \ren{multi-view images}, variable aperture \ren{pictures can be generated} after capturing. 
With this characteristic, a post-capture refocusing \ren{can also be accomplished} by properly \ren{manipulating} those light field images \cite{Levoy05}.
As light field camera provides multi-view \ren{information} from different \ren{poses}, it could help \ren{many computational photography applications}. %
As the number of views grows, more information could be retrieved.

Light field images often suffer from low spatial resolution as the sampling resources are limited. 
Cheng \albert{\etal}~\cite{DBLP:conf/cvpr/ChengXCL19} and Jin \albert{\etal}~\cite{DBLP:conf/cvpr/JinHCK20} \ren{proposed} to apply deep learning on constructing super-resolution images from light field images to cover this situation.
\ren{Depth estimation could also be achieved by leveraging the abundant information from light field images \cite{DBLP:conf/siggraph/LevoyH96,DBLP:conf/cvpr/HeberP16, DBLP:conf/iccv/Heber0P17, DBLP:conf/cvpr/DansereauGW19, DBLP:conf/aaai/TsaiLOC20, lin2015depth, zhou2019learning, park2017robust, peng2020zero}}.
Levoy \albert{\etal}~\cite{DBLP:conf/siggraph/LevoyH96} \ren{proposed} to use the epipolar plane images (EPIs), which \ren{contain both} spatial and angular information\ren{,} for depth estimation \ren{from} light field images.
Heber \albert{\etal}~\cite{DBLP:conf/cvpr/HeberP16} \ren{proposed} to learn the end-to-end mapping between the 4D light field and its corresponding 4D depth field accompanied with a high-order regularization refinement.
\ren{By following \cite{DBLP:conf/cvpr/HeberP16}, they} then designed an encoder-decoder architecture to extract the geometric information from light field images \cite{DBLP:conf/iccv/Heber0P17}.
Some works~\cite{DBLP:conf/accv/HonauerJKG16, DBLP:conf/eccv/SakurikarMBN18} focus on providing light field datasets with ground truth depth or AiF images for further use.

\subsection{Realistic Data Synthesis}

\albert{Deep learning \ren{significantly improves} the quality of computer vision tasks \ren{with a large amount of training data}.}
However, \ren{collecting} real-world data is often costly and time-consuming.
Therefore, many works \ren{instead target} synthesizing realistic data to \ren{provide sufficient data for training}.
Barron \albert{\etal}~\cite{DBLP:conf/cvpr/BarronASH15} \ren{proposed} to synthesize defocus images by a layer-wised rendering scheme \ren{with a z-buffer}. 
Wadhwa \albert{\etal}~\cite{DBLP:journals/tog/WadhwaGJFKCMBPL18} combine a \ren{person segmentation mask with a} depth map calculated from \albert{a} dual-pixel camera to generated shallow depth-of-field images.
Gur \albert{\etal}~\cite{PSFLayer} \ren{use} a differentiable point spread function, namely \ren{the} PSF convolution layer, to synthesize realistic defocus \ren{images} and train a depth from defocus network to generate depth \albert{maps}.
\yulunliu{Herrmann \ren{\etal~\cite{herrmann2020learning} proposed} a learning-based approach for autofocus and \ren{recently provided a real} dataset for training.
However, the dataset \ren{was} not open to the public at the \ren{submission time}.}

\section{Method}
\begin{figure*}
    \centering
    \includegraphics[width=\linewidth]{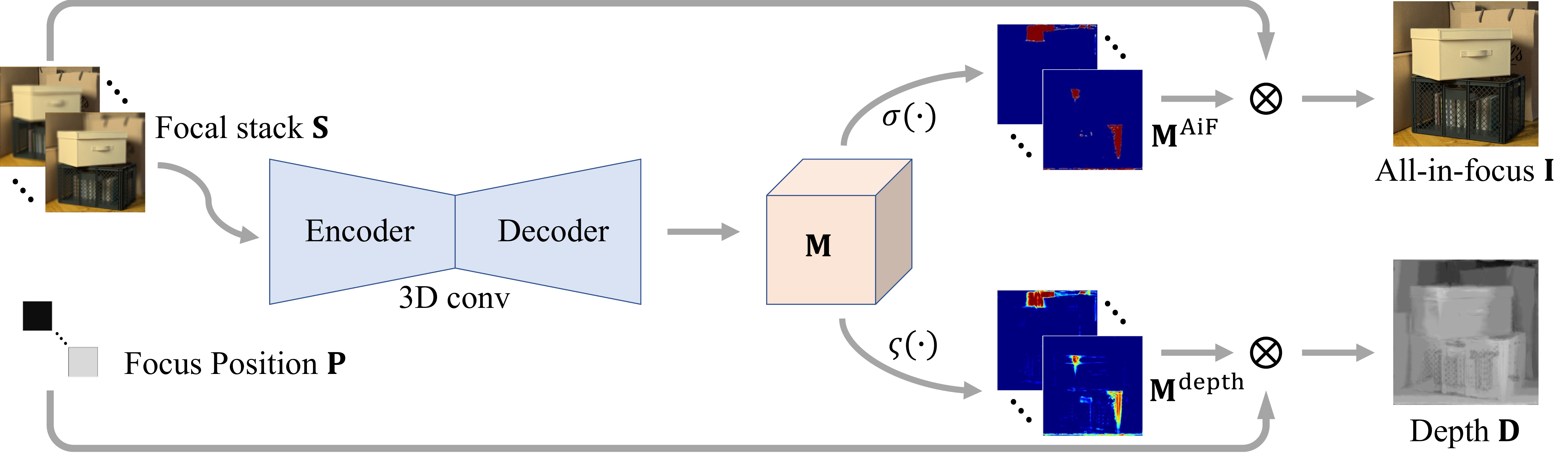}
    \caption{
        \textbf{An overview of the proposed method.}
        Given a stack of images with varying focus positions, \ie, focal stack, our model first produces an intermediate attention map $\mathbf{M}$. The intermediate attention map can be shared \ren{between} depth estimation and all-in-focus (AiF) image reconstruction. With different normalization functions, the attention map can be further manipulated to generate either depth or AiF results.
        }
    \label{fig:architecture}
\end{figure*}
We describe the proposed method in this section.
Sec.~\ref{sec:overview} gives an overview of our method. 
Then, a shared network used for \ren{both of} depth estimation and AiF image reconstruction is illustrated in Sec.~\ref{sec:network}. 
\albert{The attention mechanism that bridges the two tasks is introduced in Sec.~\ref{sec:attention}. We then depict the core concept to turn AiF images into supervisory signals for unsupervised depth estimation in Sec.~\ref{sec:unsupervise}. Finally, Sec.~\ref{sec:loss} shows the loss functions for training with depth supervision and AiF supervision.}

\subsection{Overview}\label{sec:overview}
Depth from focus aims to recover the depth from a focal stack through defocus cues. 
As shown in Fig.~\ref{fig:architecture}, 
given a focal stack $\mathbf{S} \in \mathbb{R}^{H \times W \times 3 \times F}$ of $F$ images with gradually varying focus positions $\mathbf{P} \in \mathbb{R}^{H \times W \times 1 \times F}$,
our method produces several attention representations through a shared network.
Then a depth map $\mathbf{D} \in \mathbb{R}^{H \times W \times 1}$ and an all-in-focus (AiF) image $\mathbf{I} \in \mathbb{R}^{H \times W \times 3}$ of this scene can be generated with these attention maps. 

\subsection{Network Architecture} \label{sec:network}

DDFF~\cite{DDFF} uses 2D ConvNets to address the sharpness measurement problem, which is one of the main challenges in depth from focus. 
On the other hand, DefocusNet~\cite{DefocusNet} \albert{applies} a global pooling layer as a communication tool between several weights-sharing 2D ConvNets. 
Another objective of this architecture is to allow focal stacks with arbitrary sizes. 
However, some important information across the stack dimension might not be effectively captured due to the limitation of 2D convolution and simple global pooling.

For this reason, we adopt the Inception3D~\cite{I3D} as the backbone of our model. 
As shown in Fig.~\ref{fig:architecture}, our model is an encoder-decoder network consisting of 3D convolutions. 
With the 3D convolution, defocus cues could be better captured across frames and thus facilitate the tasks of depth estimation and AiF image reconstruction. 
Moreover, our model can also handle focal stacks with %
\albert{arbitrary sizes} attributed to the nature of 3D convolution.

\subsection{Attention Mechanism} \label{sec:attention}

The output of our network is an intermediate attention $\mathbf{M}  \in \mathbb{R}^{H \times W \times 1 \times F}$.
The underlying expectation of the intermediate attention $\mathbf{M}$ is that it should reflect the probability of each focus position leading to the maximal sharpness. 
Then it can benefit \ren{both} of depth estimation and AiF image reconstruction.

For depth estimation, we propose to normalize the intermediate attention $\mathbf{M}$ into a depth attention $\mathbf{M}^{\text{depth}}$ via a softplus normalization: 
\begin{align} \label{eq:softplus}
    \mathbf{M}^{\text{depth}} = \varsigma(\mathbf{M}),
\end{align}
where
\begin{align} \label{eq:softplus}
    \mathbf{M}^{\text{depth}}_{i, j, 1, t} = \frac{\ln{(1 + \exp{(\mathbf{M}_{i, j, 1, t})})}}{\sum_{n=1}^{F}\ln{(1 + \exp{(\mathbf{M}_{i, j, 1, n})})}}.
\end{align}
The softplus function is a smooth version of ReLU.

The depth attention $\mathbf{M}^{\text{depth}}$ can also be interpreted as the probability distribution because the softplus function ensures non-negativeness and normalizes $\mathbf{M}^{\text{depth}}$ into a valid probability distribution.
Then the expected depth value of each pixel can be derived via:
\begin{equation}
\mathbf{D}_{i, j, 1} = \Sigma_{t=1}^{F}(\mathbf{M}^{\text{depth}} \cdot \mathbf{P})_{i, j, 1, t}.
\end{equation}

For AiF image reconstruction, we perform similar procedures except that the normalization function changes from softplus to softmax.
That is, the AiF attention $\mathbf{M}^{\text{AiF}}$ is obtained as: \begin{align} \label{eq:softmax}
    \mathbf{M}^{\text{AiF}} = \sigma(\mathbf{M}),
\end{align}
where
\begin{align} \label{eq:softmax}
    \mathbf{M}^{\text{AiF}}_{i, j, 1, t} = \frac{\exp{(\mathbf{M}_{i, j, 1, t})}}{\sum_{n=1}^{F}{\exp{(\mathbf{M}_{i, j, 1, n})}}}.
\end{align}

The AiF attention $\mathbf{M}^{\text{AiF}}$ can then be used for AiF image reconstruction. The expected AiF image of each pixel is expressed as:
\begin{equation}
\mathbf{I}_{i, j, k} = \Sigma_{t=1}^{F}(\mathbf{M}^{\text{AiF}} \cdot \mathbf{S})_{i, j, k, t}.
\end{equation}

The reason why we adopt \ren{the} softmax and softplus normalization\ren{s} separately for depth and AiF image estimation is to tackle the problem of \emph{sparse} focal stacks.
\albert{By sparse, we mean that the stack size is small, and the focus positions inside the focal stack are not dense.}

For AiF image reconstruction, it is best to select the sharpest pixel along the stack dimension.  
Blending multiple pixels inside a sparse stack usually does not help.
\albert{Therefore, we leverage the softmax function to pursue the peaking phenomenon to extract the clearest pixels.}

\begin{figure}[]
    \centering
    \renewcommand{\tabcolsep}{1pt} %
    \renewcommand{\arraystretch}{1} %
    \newcommand{\imagewidth}{0.47\columnwidth}
    \begin{tabular}{ccc}
        \\
        &{\small Supervised} & {\small Unsupervised }\\
        \raisebox{2.5\normalbaselineskip}[0pt][0pt]{
            \rotatebox[origin=c]{90}{\footnotesize Same}} &
        \includegraphics[width=\imagewidth]{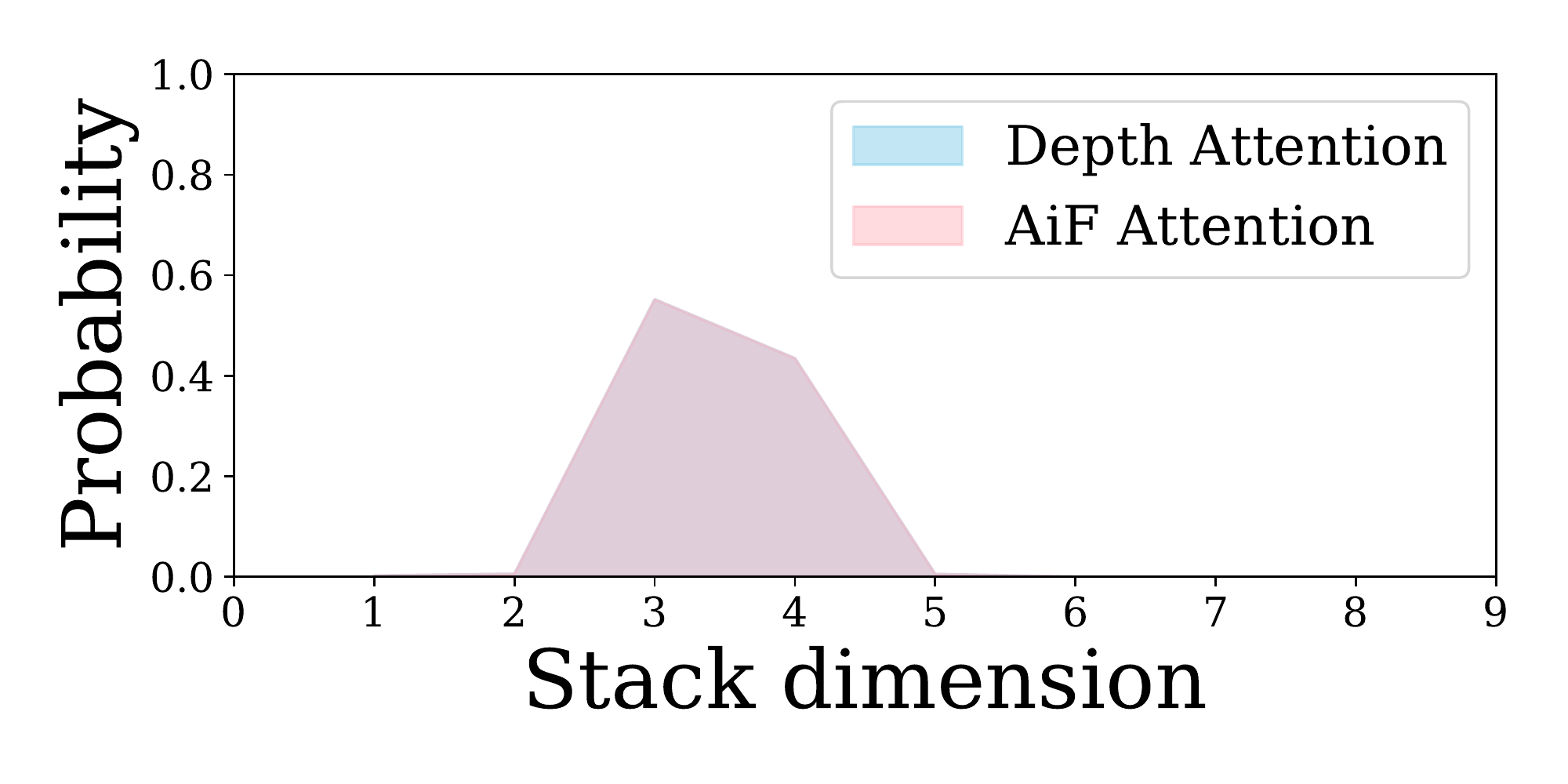} &
        \includegraphics[width=\imagewidth]{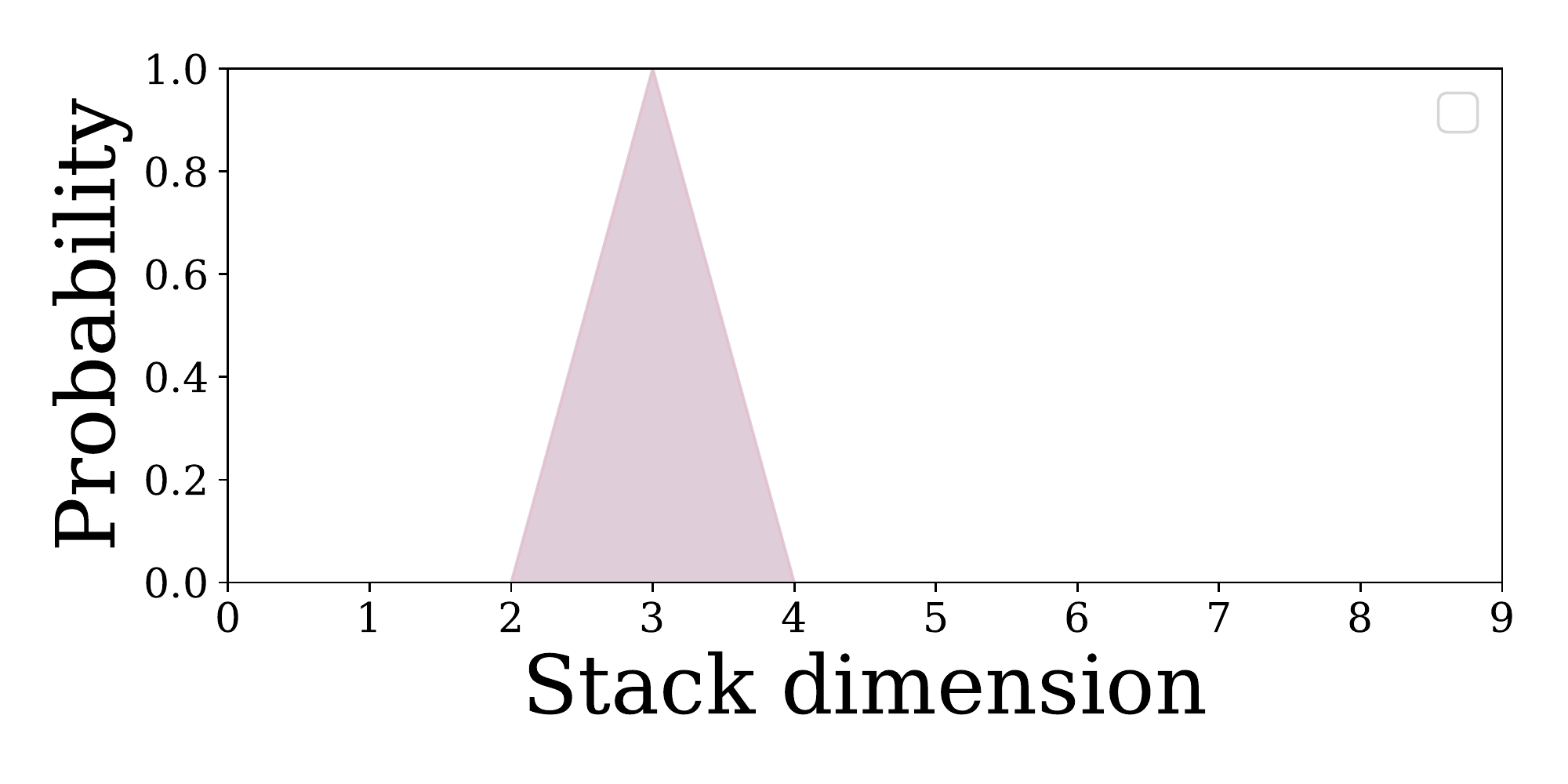}\\
        \raisebox{2.5\normalbaselineskip}[0pt][0pt]{\rotatebox[origin=c]{90}{\footnotesize Seperate}} &
        \includegraphics[width=\imagewidth]{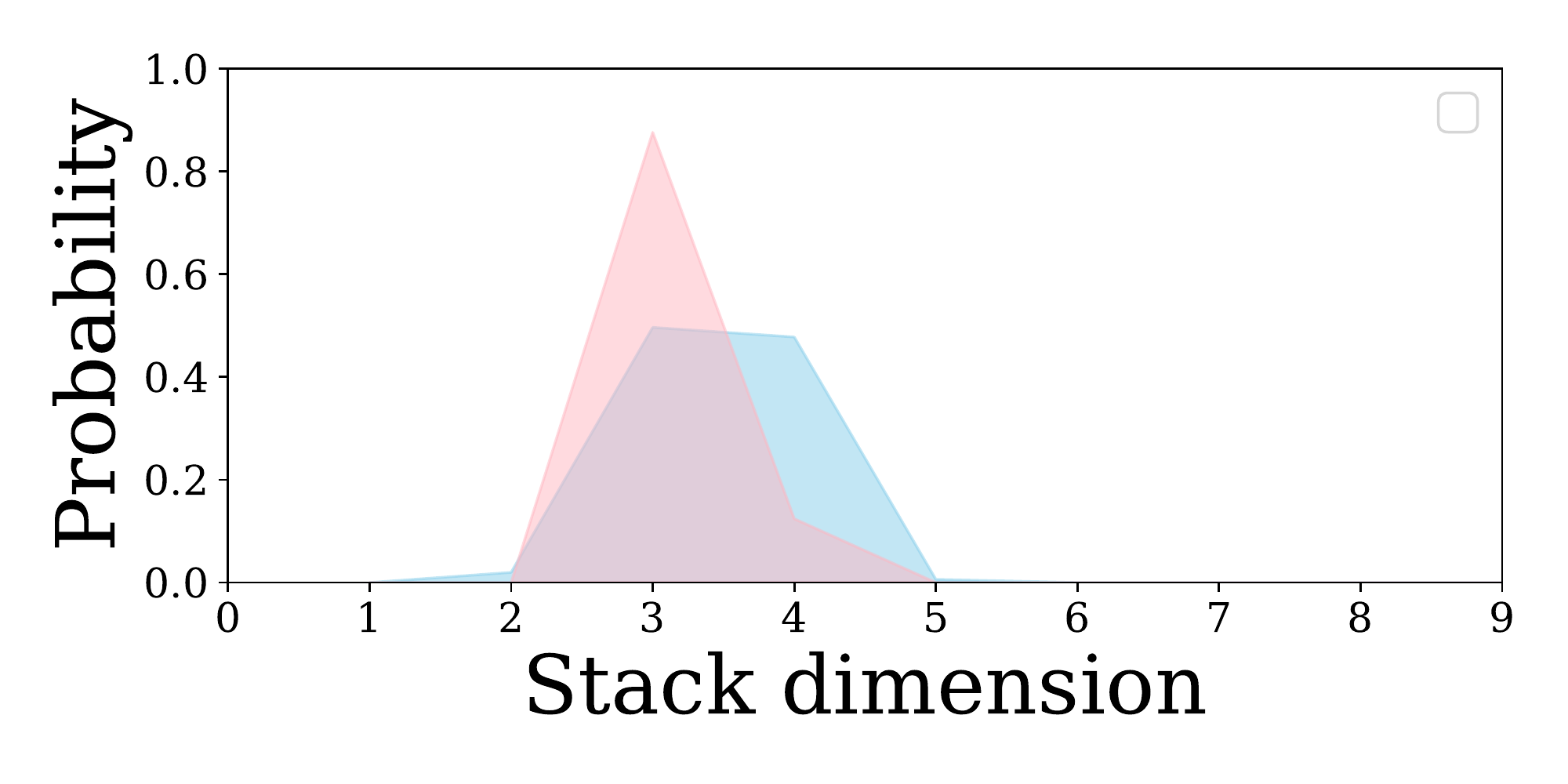} &
        \includegraphics[width=\imagewidth]{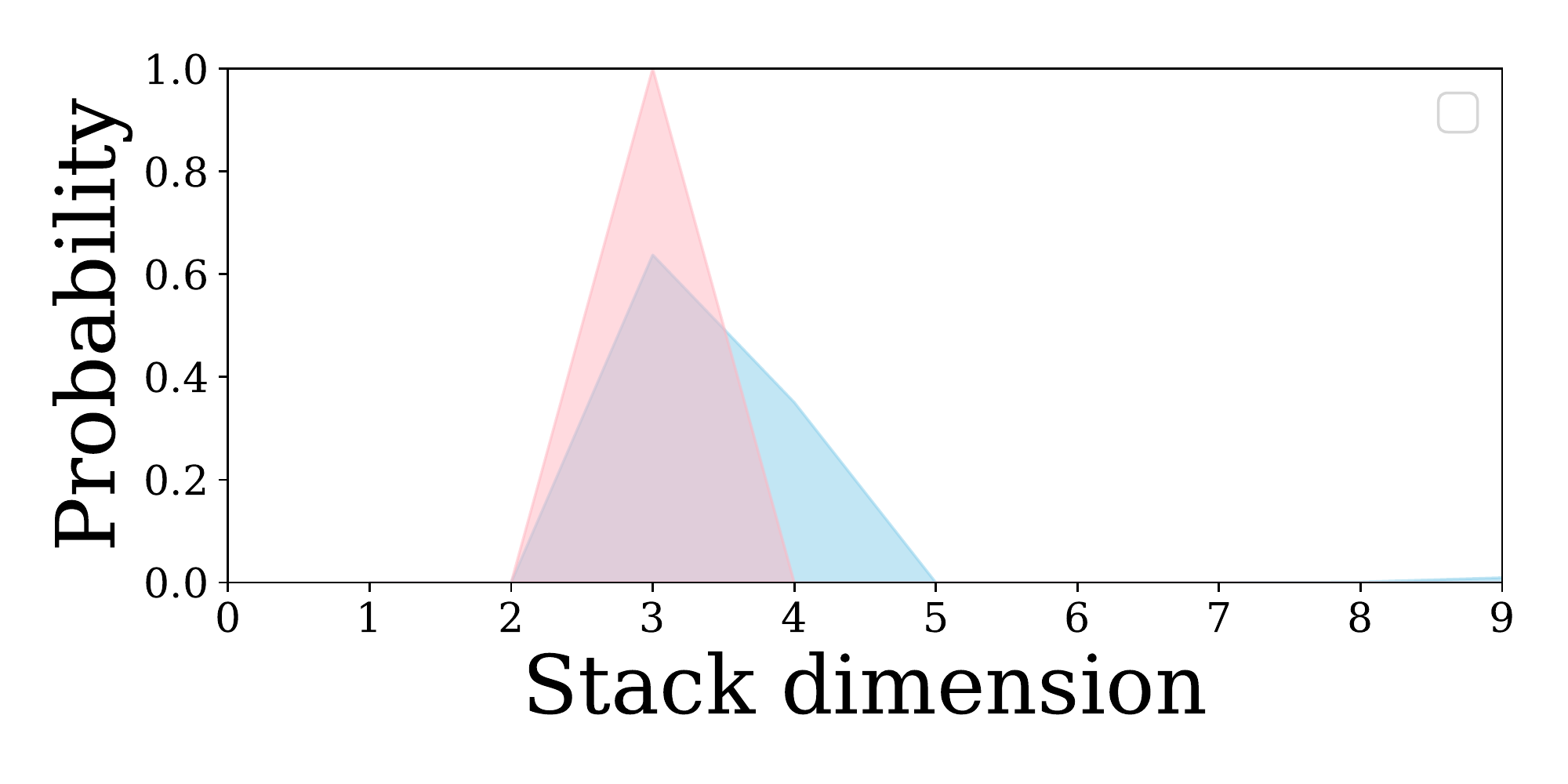} \\
    \end{tabular}%
    \caption{\textbf{The effect of normalization on attention maps.}
    The first row shows that if we adopt the same softmax normalization for depth and AiF attention, both of them tend to be a flatter distribution in supervised learning and face a peaking phenomenon in unsupervised learning \ren{(AiF supervision)}. The second row shows that with separate normalization as described in Sec.~\ref{sec:attention}, each of depth and AiF attention becomes a proper distribution in both supervised and unsupervised learning.
    }
    \label{fig:AiF_Disp_Attention_ablation}
\end{figure}

For depth estimation, \ren{the} softmax normalization results in severe quantization for sparse focal stacks because it simply selects the nearest focal position of maximal sharpness.
On the other hand, \ren{the} softplus normalization leads to a flatter distribution so that more accurate depth can be predicted by interpolating among sparse focal positions. 
Fig.~\ref{fig:AiF_Disp_Attention_ablation} shows the effects of this design choice.

\subsection{Towards Unsupervised Depth Estimation}\label{sec:unsupervise}

Similar to our method, DefocusNet~\cite{DefocusNet} also combines the tasks of depth estimation and AiF image reconstruction via \emph{intermediate defocus maps}. 
Their intermediate defocus maps necessitate supervisory signals, which can be derived by calculating \albert{the} circle of confusions from the ground truth depth. 
Furthermore, they also propose to use a DepthNet following the intermediate defocus maps to predict the output depth. 
Therefore, their method can only be trained supervisedly when ground truth depth data are available.

Instead, we propose to use the intermediate attention $\mathbf{M}$ to bridge the two tasks, which does not necessitate intermediate supervisory signals \ren{from ground truth depth data}.
Moreover, there \ren{are} no learnable parameters after $\mathbf{M}$.
We only use fixed normalization functions to generate the output depth and AiF images.
Therefore, even when ground truth depth data are not available, we can still train the shared network to generate the intermediate attention $\mathbf{M}$ only via supervisory signals from ground truth AiF images.
That is, our method can be trained supervisedly or unsupvervisedly with or without ground truth depth data.

\subsection{Training Loss}\label{sec:loss}

For supervised depth estimation , our model is trained with a simple $L_{1}$ loss:
\begin{align} \label{eq:supervise}
    L_\text{supervised} = L_\text{depth} = \mathbb{E} [ \| \mathbf{D} - \mathbf{D}_\text{gt} \|_{1} ],
\end{align}
where $\mathbf{D}_\text{gt}$ stands for the ground truth depth.

For unsupervised depth estimation, our model can also be trained by a $L_1$ loss with AiF supervison:
\begin{align} \label{eq:aif}
    L_\text{AiF} = \mathbb{E} [ \| \mathbf{I} - \mathbf{I}_\text{gt} \|_{1} ],
\end{align}
where $\mathbf{I}_\text{gt}$ denotes the ground truth AiF image. 
Furthermore, we also encourage our depth map to be locally smooth using an edge-aware weighting as in~\cite{MonoDepth}. The smoothness loss is defined as:
\begin{align} \label{eq:smooth}
    L_\text{smooth} = \mathbb{E} [ W_{x} \left| \frac{\partial{\mathbf{D}_{i, j, 1}}}{\partial{x}} \right| + W_{y} \left| \frac{\partial{\mathbf{D}_{i, j, 1}}}{\partial{y}} \right| ],
\end{align}
where
\begin{equation}
\begin{split}
    W_x = \exp{(-\frac{\lambda}{3}\sum_{k}{\left| \frac{\partial{\mathbf{I}_{i, j, k}}}{\partial{x}} \right|})}
    \\
    W_y = \exp{(-\frac{\lambda}{3}\sum_{k}{\left| \frac{\partial{\mathbf{I}_{i, j, k}}}{\partial{y}} \right|})}, 
\end{split}
\end{equation}
and $\lambda$ is a hyper-parameter for the edge weighting based on the ground truth AiF image.
The total loss of our unsupervised depth estimation is then: 
\begin{align} \label{eq:unsupervise}
    L_\text{unsupervised} = L_\text{AiF} + \alpha L_\text{smooth},
\end{align}
where $\alpha$ indicates the importance of the smoothness loss.

\section{Evaluation}
In this section, we describe the datasets used in the experiments, and report ablation studies and comparisons with the state-of-the-art methods on depth from focus. %
In the supplementary material, we present additional results for AiF image reconstruction, \ren{and visual comparisons on DDFF 12-Scene}~\cite{DDFF}, \ren{Middlebury Stereo Datasets}~\cite{Middlebury}, and the \ren{DefocusNet} dataset~\cite{DefocusNet}.

\begin{table*}[h]
    \centering
    \small
    \caption{\textbf{\ren{Summary of the evaluation datasets}.}
    }
    \label{tab:dataset_summarize}
    \resizebox{\textwidth}{!}{%
    \begin{tabular}{l|cccc}
        \toprule
        Dataset & Image source & Cause of defocus & Disparity/Depth GT & AiF GT \\
        \midrule
        \ren{DefocusNet}~\cite{DefocusNet} & Synthetic & Blender rendering & Depth & \\
        \ren{DDFF 12-Scene}~\cite{DDFF} & Real & Light-field composition & Depth &  \\
        \ren{4D Light Field Dataset}~\cite{honauer2016dataset} & Synthetic & Light-field composition & Disparity & \checkmark \\
        \ren{Middlebury Stereo Datasets}~\cite{Middlebury} & Real & Disparity rendering & Disparity & \checkmark \\
        \ren{Mobile Depth}~\cite{MobilePhone} & Real & Real &  & Stitching by MRF  \\
        \bottomrule
    \end{tabular}
    }
\end{table*}

\subsection{Datasets}

Totally five datasets are used in the quantitative experiments and visual comparisons. 
Their descriptions and statistics are summarized in \tabref{dataset_summarize}. For more details, please see the supplementary material.

\subsection{Evaluation Metrics}
In this paper, we evaluate quantitative results with the following metrics\ren{:} mean-absolute error (MAE), mean-squared error (MSE), root-mean-squared error (RMSE), log root-mean-squared error (logRMS), relative-absolute error (Abs. rel.), relative-squared error (Sqr. rel.), bumpiness (Bump), accuracy with $\delta=1.25$, per-stack inference time (Secs.).

\subsection{Implementation Details}
We implement our method in PyTorch~\cite{Pytorch}.
Our model is trained from scratch using the Adam optimizer~\cite{Adam} ($\beta_1 = 0.9$, $\beta_2 = 0.999$), with a learning rate of $10^{-4}$.
For \ren{DDFF 12-Scene}~\cite{DDFF}, \ren{4D Light Field Dataset}~\cite{DBLP:conf/accv/HonauerJKG16}, \ren{FlyingThings3D}~\cite{FlyingThings3D}, and \ren{the DefocusNet  dataset}~\cite{DefocusNet}, we use 10, 10, 15, and 5 as the input stack sizes, respectively.
In \ren{unsupervised learning}, the weight of smoothness loss $\alpha$ is set to $0.002$ in all experiments.
We apply random spatial transformations (flipping, cropping, rotation) and random color jittering (brightness, contrast, and gamma) for data augmentation during training. 
The training patch size is $256 \times 256$ after random cropping except for \ren{DDFF 12-Scene}~\cite{DDFF}.
\ren{Because the} image size of \ren{DDFF 12-Scene} is $224 \times 224$, \ren{which is larger than $256 \times 256$,} we do not apply random cropping on this dataset.
All the experiments are conducted with a single NVIDIA GTX 1080 GPU.

\subsection{Ablation Studies}
To understand the performance of \ren{each} proposed components in our method, we conduct various ablation studies on \ren{4D Light Field Dataset}~\cite{honauer2016dataset}. \ren{All models are supervised trained with disparity maps if not specified}.

\heading{Architecture.}
Table~\ref{table:Architecture_ablation} shows the \ren{comparison} between 2D and 3D convolutions and with/without the proposed attention \ren{mechanism}.
The 3D convolution performs better than 2D convolution as the 3D convolution is able to capture and aggregate features across both spatial and stack dimensions.
The design of attention bridges the tasks of depth estimation and AiF image reconstruction\albert{, \ren{and thus} enables the possibility of unsupervised depth estimation as described in Sec.~\ref{sec:unsupervise} while maintaining roughly the same prediction quality.}

\begin{table}[]
    \centering
    \small
    \caption{
    \textbf{Ablation study on network architecture.}
    The 3D convolution leads to better performance on disparity estimation because of its ability to capture features for both spatial and stack dimensions.
    Although the design of attention does not increase the performance explicitly, it bridges the tasks of depth estimation and AiF image reconstruction and thus enables the unsupervised learning for depth estimation.
        }
        \resizebox{\columnwidth}{!}{%
    \begin{tabular}{c|c|cccccc}
        \toprule
        Architecture & Attention &
        \multicolumn{1}{c}{{MAE}$\downarrow$} &
        \multicolumn{1}{c}{{MSE}$\downarrow$} &
        \multicolumn{1}{c}{{RMSE}$\downarrow$} & 
        \multicolumn{1}{c}{{Bump.}$\downarrow$} \\ 
        \midrule
        \multirow{2}{*}{3D} & \checkmark &
        0.0788 &
        0.0472 &
        0.2014 &
        1.5776 
        \\
         &  &
        0.0851 & 
        0.0461 & 
        0.1984 & 
        2.3168 
        \\
        \midrule
        \multirow{2}{*}{2D} &\checkmark &
        0.1070 &
        0.0577 &
        0.2259 &
        2.1435 
        \\
         &  &
        0.1179 &
        0.0576 &
        0.2291 &
        2.3168 
        \\
        \bottomrule
    \end{tabular}
    }
    \label{table:Architecture_ablation}
\end{table}

\begin{figure}
    \centering
    \includegraphics[width=\linewidth]{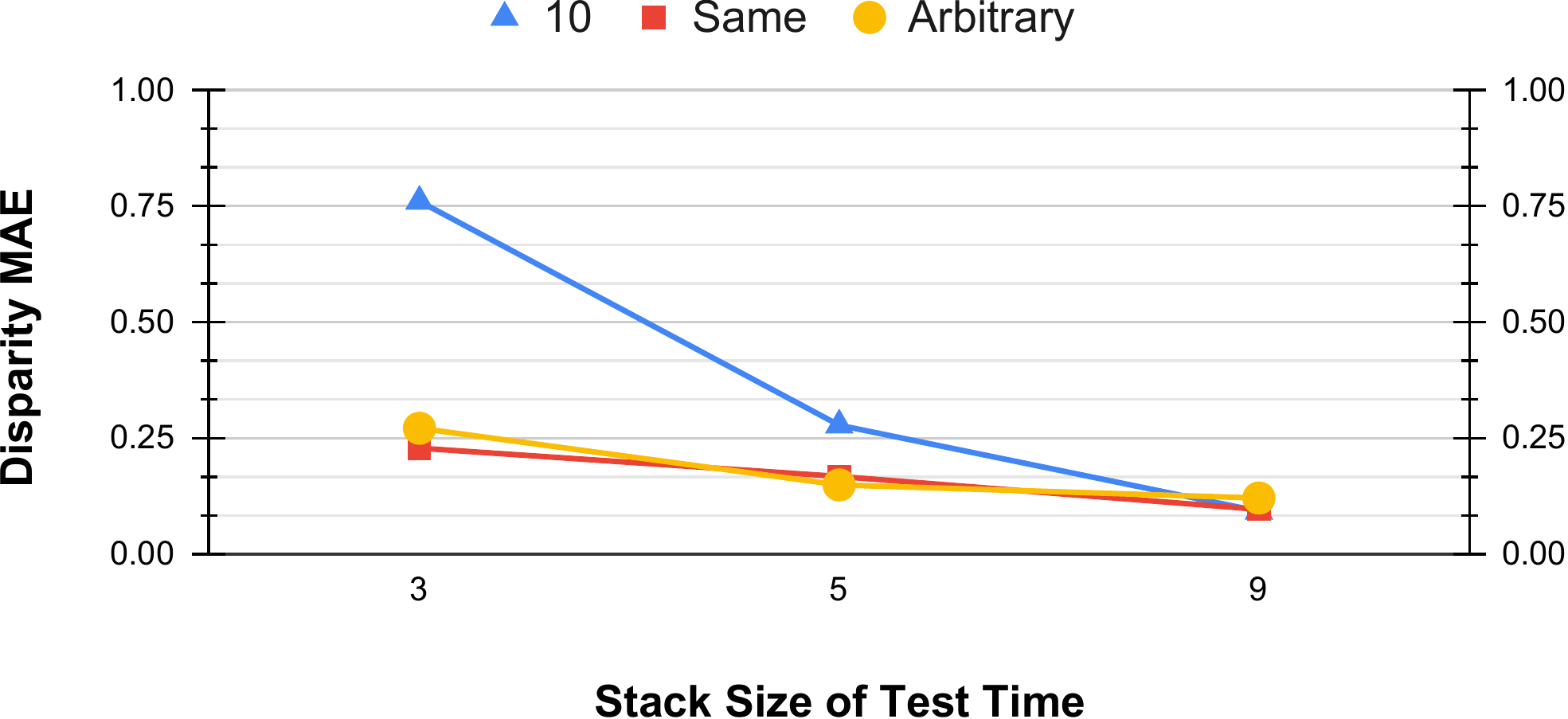}
    \caption{
        \textbf{The effect of focal stack size.}
        Each line indicates a stack size setting during training. Specifically, \emph{10} and \emph{Arbitrary} respectively represent training with a fixed size of $10$ and training with arbitrary sizes, and \emph{Same} refers to using the same setting as test stacks.
    }
    \label{fig:stack_num_line}
\end{figure}

\heading{Focal stack size.}
Our network can handle arbitrary input stack \ren{sizes} due to the nature of 3D convolution. 
Fig.~\ref{fig:stack_num_line} shows a comparison on different input stack sizes.
We train the \ren{models} in three different ways: 1) \emph{10}: training with a \emph{fixed} input stack size of 10. 2) \emph{Arbitrary}: training with \emph{arbitrary} input stack sizes randomly sampled from 2 to 10. 3) \emph{Same}: training with the \emph{same} size as test stacks.

The model trained with a fixed stack size performs poorly while testing with different stack sizes.
As expected, the model trained with the same stack size as test data performs the best as the input setting for training and testing are consistent.
However, the model \ren{trained} with arbitrary input stack sizes performs favorably against the ones \ren{trained} %
with the same stack size as test data.
This demonstrates the robustness of our method across different input stack sizes.%

\begin{figure}[]
    \centering
    \renewcommand{\tabcolsep}{1pt} %
    \renewcommand{\arraystretch}{1} %
    \newcommand{\imagewidth}{0.185\columnwidth}
    \begin{tabular}{ccccc}
            \includegraphics[width=\imagewidth]{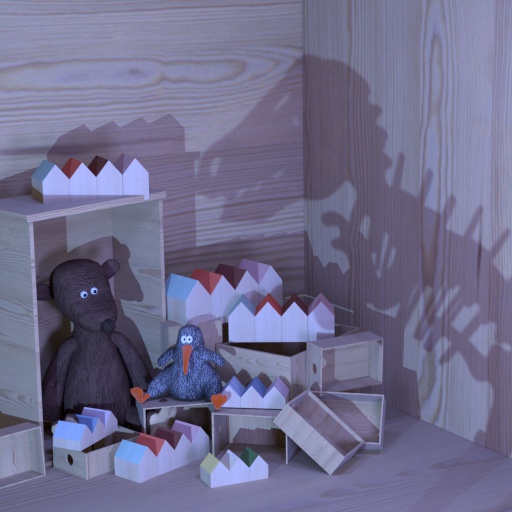} &
            \includegraphics[width=\imagewidth]{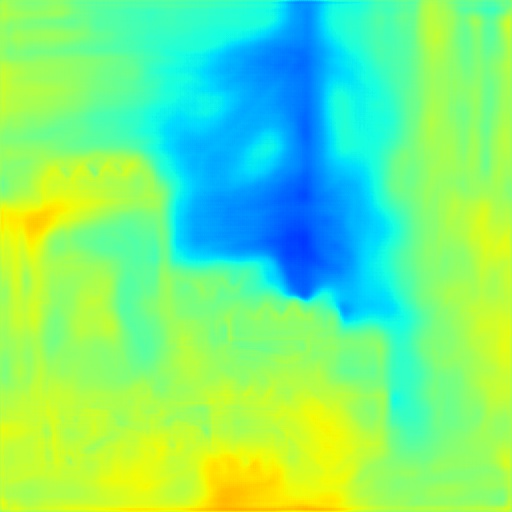} & 
            \includegraphics[width=\imagewidth]{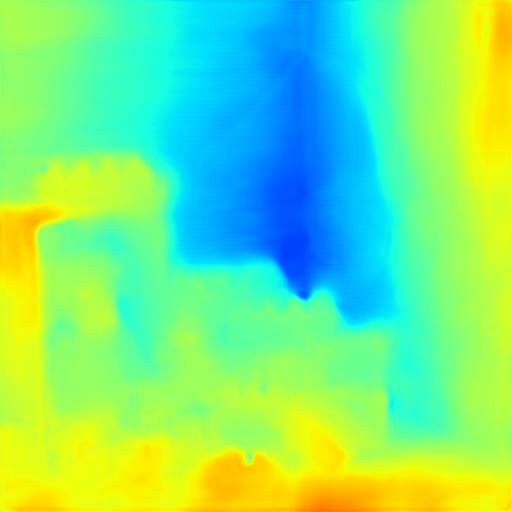} & 
            \includegraphics[width=\imagewidth]{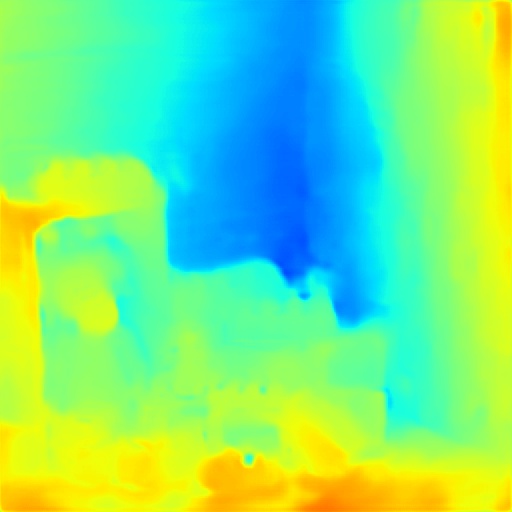} & 
            \includegraphics[width=\imagewidth]{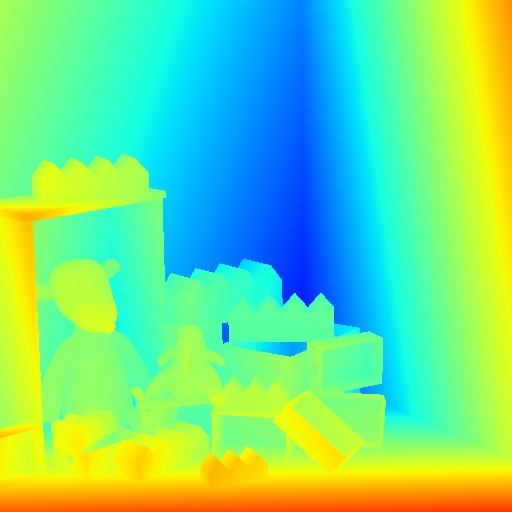} \\
            RGB & stack size 3 & stack size 5 & stack size 9 & GT \\
    \end{tabular}%
    \caption{
    \textbf{Visual comparison on different focal stack sizes.}
    The quality of the estimated disparity map improves as the stack size increases.
    }
    \label{fig:HCI_stacksize_ablation}
\end{figure}

Fig.~\ref{fig:HCI_stacksize_ablation} shows the visual comparison of different input stack sizes, generated with the model trained with arbitrary input stack sizes.
It is obvious that the quality of the estimated disparity map improves as the stack size increases.

\begin{figure}[]
    \centering
    \footnotesize
    \renewcommand{\tabcolsep}{1pt} %
    \renewcommand{\arraystretch}{1} %
    \newcommand{\imagewidth}{0.15\columnwidth}
    \resizebox{\columnwidth}{!}{%
    \begin{tabular}{cccccc}
            \includegraphics[width=\imagewidth]{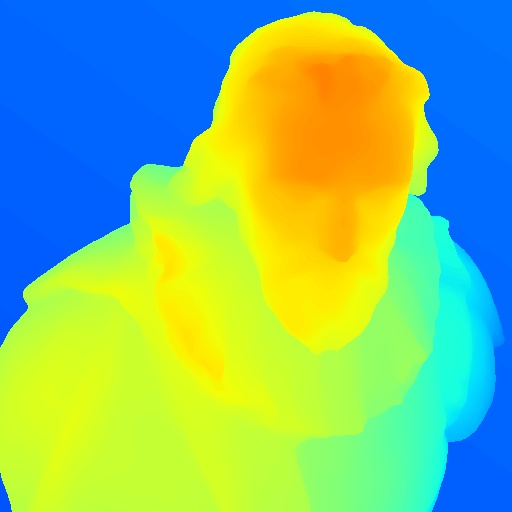} & 
            \includegraphics[width=\imagewidth]{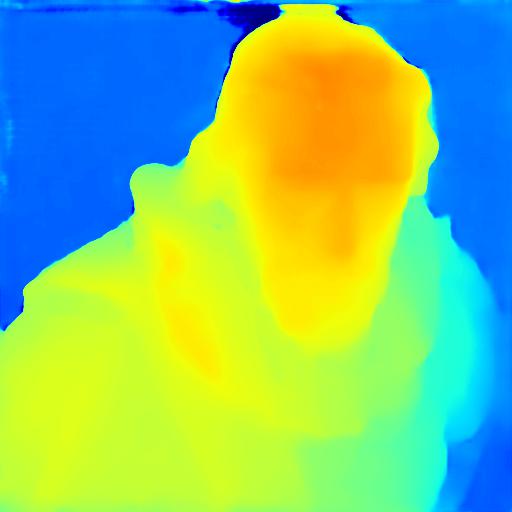} &
            \includegraphics[width=\imagewidth]{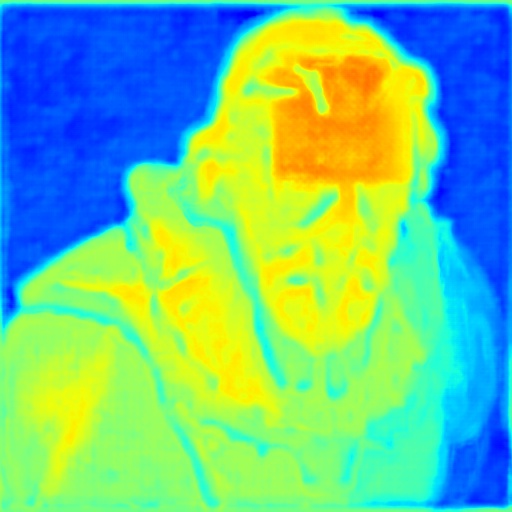} &
            \includegraphics[width=\imagewidth]{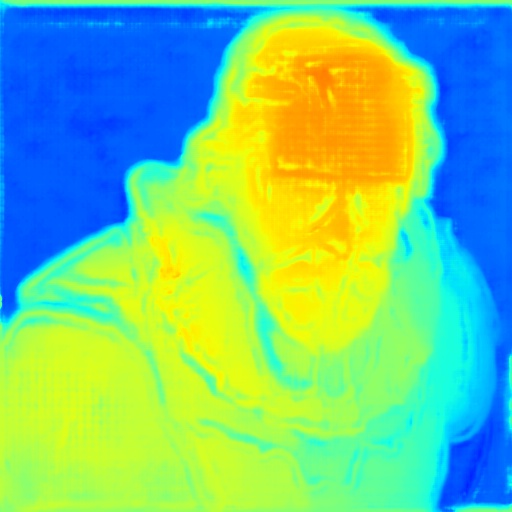} &
            \includegraphics[width=\imagewidth]{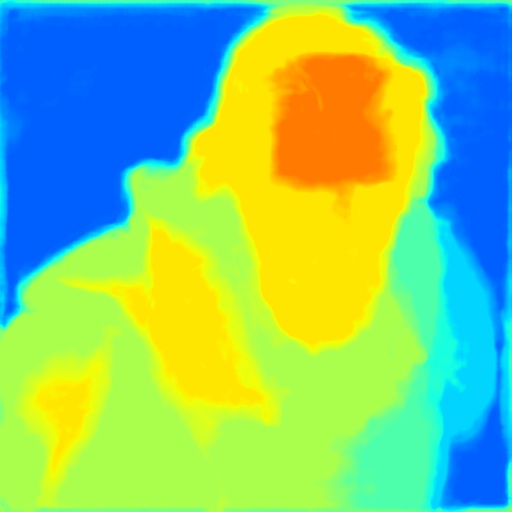} & 
            \includegraphics[width=\imagewidth]{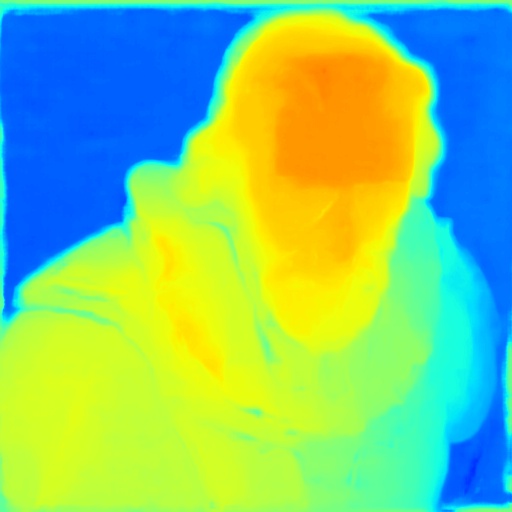}
            \\
            GT &  \\
            \ren{Supervised} & Yes & No & No & No & No \\
            Stack size &  & 10 & 37 & 10 & 37 \\
            $L_{\text{smooth}}$ &  &  &  & \checkmark & \checkmark \\
            
    \end{tabular}%
    }
    \caption{\textbf{Visual comparison on different supervision settings.}
    The results of supervised \ren{learning} %
    \ren{are} better than the ones from unsupervised \ren{learning (AiF supervision)}. %
    The \ren{output} disparity maps from unsupervised \ren{learning} %
    suffer from \ren{the} quantization effect. By adding the smoothness loss, the disparity results %
    become locally smooth and perform favorably against the ones from supervised \ren{learning} qualitatively.
    }
    \label{fig:AiF_ablation}
\end{figure}

\heading{Unsupervised \ren{learning} using all-in-focus images.}
\ren{On the basis of the attention mechanism} described in Sec.~\ref{sec:attention}, our method is able to \ren{be trained with either ground truth disparity maps (supervised) or ground truth AiF images (unsupervised)}.
\ren{Table~\ref{table:AiF_ablation} and Fig.~\ref{fig:AiF_ablation} respectively show the quantitative and qualitative results for both of our supervised and unsupervised models}.
As shown in Table~\ref{table:AiF_ablation}, with a large input stack size of $37$ and the smoothness loss, the performance of unsupervised \ren{learning} is able to approximate that of supervised \ren{learning} with a smaller stack size of $10$ .

However, the results of \ren{unsupervised learning} with a smaller stack size suffer from the quantization problem as described in Sec.~\ref{sec:unsupervise}. \ren{As shown in Fig.~\ref{fig:AiF_ablation}, after} adding the smoothness loss, the output disparity \ren{map becomes} locally smooth and perform better \ren{qualitatively}.

\begin{table}[]
    \centering
    \small
    \caption{\textbf{Ablation study on supervision.} 
    The results of supervised \ren{learning} perform better than the ones from unsupervised \ren{learning (AiF supervision)}. 
    Unsupervised \ren{learning} often generate\ren{s} disparity maps \ren{that} suffer from \ren{the} quantization effect and lead to poor results.
    After adding the smoothness loss, the output disparity maps become locally smooth and perform better \ren{quantitatively}. %
        }
    \resizebox{\columnwidth}{!}{%
    \begin{tabular}{ccc|ccc}
        \toprule
        Supervised & Stack size & $L_{smooth}$ &
        \multicolumn{1}{c}{{MAE}$\downarrow$} &
        \multicolumn{1}{c}{{MSE}$\downarrow$} &
         \multicolumn{1}{c}{{RMSE}$\downarrow$} 
        \\ 
        \midrule
        Yes & 10 &  &
        0.0788 &
        0.0472 &
        0.2014 
        \\
        No & 10 &  &
        0.2425 &
        0.1174 &
        0.3401 
        \\
        No & 37 &  &
        0.2099 &
        0.1039 &
        0.3202 
        \\
        No & 10 & \checkmark &
        0.1671 &
        0.0746 &
        0.2698 
        \\
        No & 37 & \checkmark &
        0.1116 &
        0.0584 &
        0.2311 
        \\
        \bottomrule
    \end{tabular}
    }
    \label{table:AiF_ablation}
\end{table}

\subsection{Comparisons to the State-of-the-art Methods}

After ablation on different components of the proposed method, we conduct comparisons with \ren{the} state-of-the-art methods on \ren{various} datasets in this section.

\heading{\ren{DDFF 12-Scene}.}
Table~\ref{table:DDFF_baseline} shows the quantitative \ren{comparison} on \ren{DDFF 12-Scene}~\cite{DDFF}.
All methods are supervisedly trained on this dataset with ground truth depth.
The proposed method performs favorably against the state-of-the-art methods under all metrics.
Furthermore, our method runs faster than \ren{another deep learning based method DDFF~\cite{DDFF}} . %

\begin{table*}[]
    \center
    \small
    \caption{
    \textbf{Quantitative comparison on \ren{DDFF 12-Scene}.} Note that \ren{RMSE is not presented} due to the mismatched results \ren{from the benchmark website and the paper of DDFF~\cite{DDFF}}. \ren{For DefocusNet~\cite{DefocusNet}, we only show the original metric reported in their paper, which is MSE.} \best{Red} text indicates the best, and \second{blue} text indicates the second-best performing method. }
    \begin{tabular}{l|cccccccc}
        \toprule
        Method & MSE $\downarrow$ & log RMS $\downarrow$ &
        Abs. rel. $\downarrow$ &
        Sqr. rel. $\downarrow$ &
        Bump. $\downarrow$ &
        $\delta=1.25$ $\uparrow$ &
        $\delta=1.25^2$ $\uparrow$ &
        $\delta=1.25^3$ $\uparrow$ \\
        \midrule
        Ours & \best{$8.6e^{-4}$} & \best{0.29} & \best{0.25} & \best{0.01} & 0.63 & \best{68.33} & \best{87.40} & \second{93.96} \\
        \ren{DefocusNet}~\cite{DefocusNet} & \second{$9.1e^{-4}$} & - & - & - & - & - & - & - \\
        DDFF~\cite{DDFF} & $9.7e^{-4}$ & 0.32 & 0.29 & \best{0.01} & 0.59 & 61.95 &      85.14 & 92.98 \\
        PSPNet~\cite{PSPNet}  & $9.4e^{-4}$ & \best{0.29} & 0.27 & \best{0.01} & \second{0.55}     & \second{62.66} & \second{85.90} & \best{94.42} \\
        Lytro & $2.1e^{-3}$ & \second{0.31} & \second{0.26} & \best{0.01} & 1.02  & 55.65 &      82.00 & 93.09 \\
        PSP-LF~\cite{PSPNet} & $2.7e^{-3}$ & 0.45 & 0.46 & \second{0.03} & \best{0.54}  &     39.70 & 65.56 & 82.46 \\
        DFLF~\cite{DDFF} & $4.8e^{-3}$ & 0.59 & 0.72 & 0.07 & 0.65  & 28.64 &       53.55 & 71.61 \\
        VDFF~\cite{VDFF} & $7.3e^{-3}$ & 1.39 & 0.62 & 0.05  & 0.79 & 8.42 &        19.95 & 32.68 \\
        \bottomrule
    \end{tabular}
    \label{table:DDFF_baseline}
\end{table*}

\begin{figure*}[]
    \centering
    \footnotesize
    \renewcommand{\tabcolsep}{1pt} %
    \renewcommand{\arraystretch}{1} %
    \newcommand{\imagewidth}{0.25\columnwidth}
    \begin{tabular}{cccccccc}
    \includegraphics[width=\imagewidth]{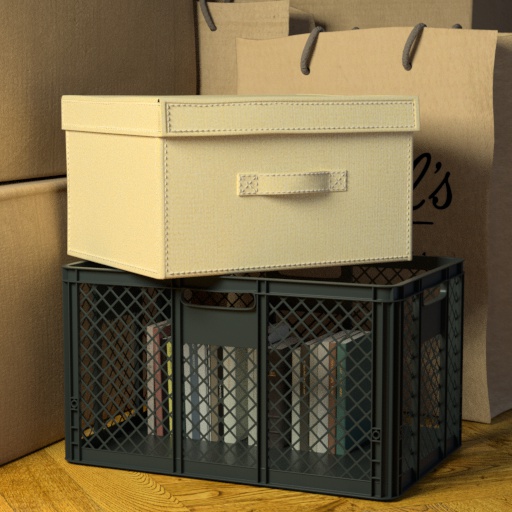} &
    \includegraphics[width=\imagewidth]{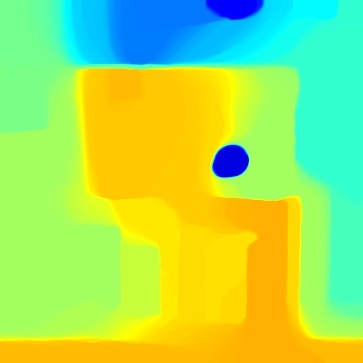} &
    \includegraphics[width=\imagewidth]{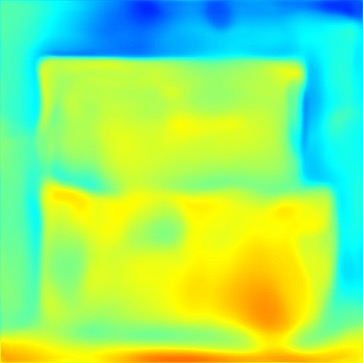} &
    \includegraphics[width=\imagewidth]{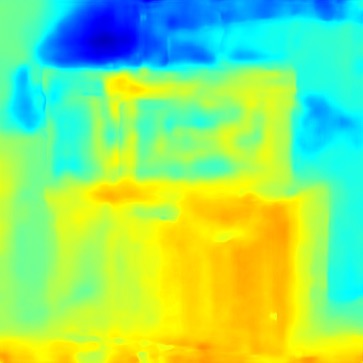} &
    \includegraphics[width=\imagewidth]{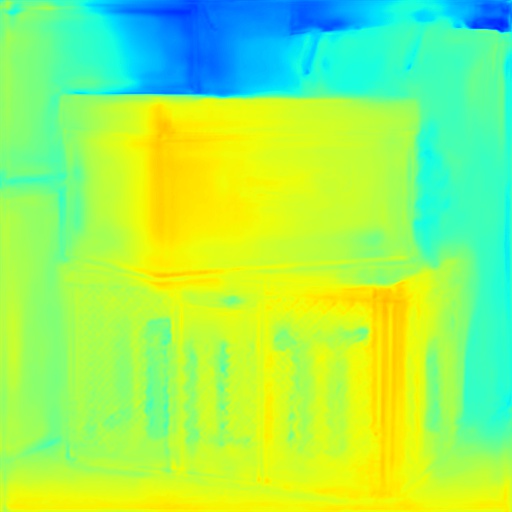} &
    \includegraphics[width=\imagewidth]{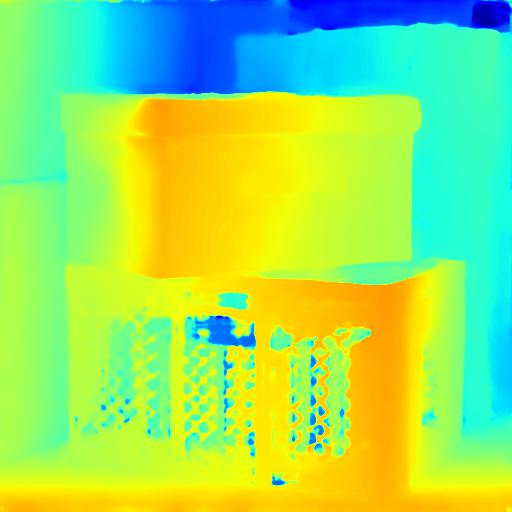} &
    \includegraphics[width=\imagewidth]{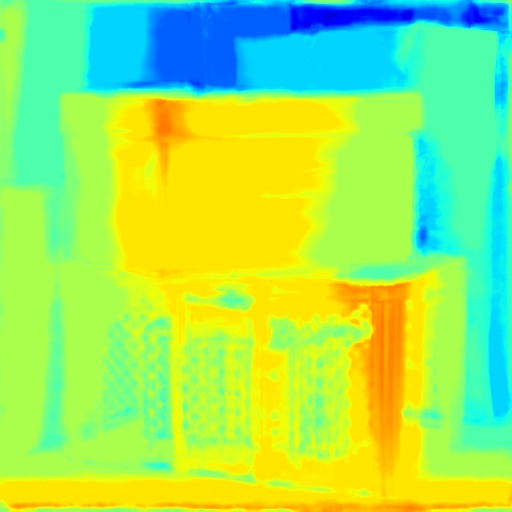} & 
    \includegraphics[width=\imagewidth]{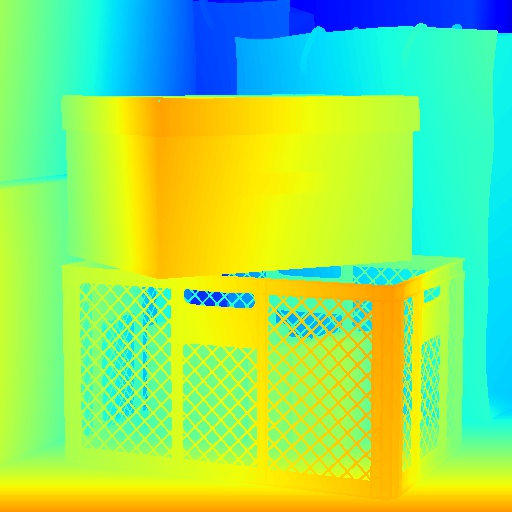}
    \\
    \includegraphics[width=\imagewidth]{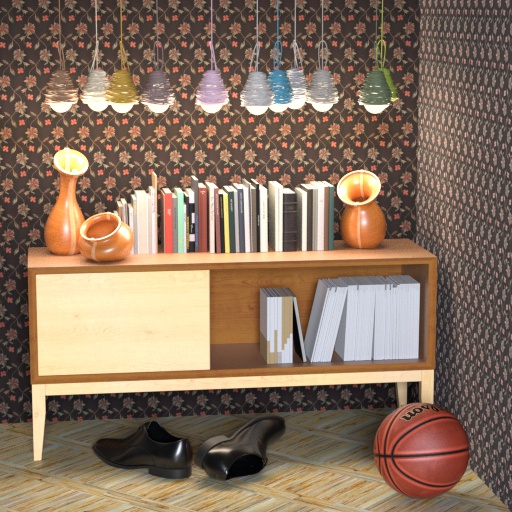} &
    \includegraphics[width=\imagewidth]{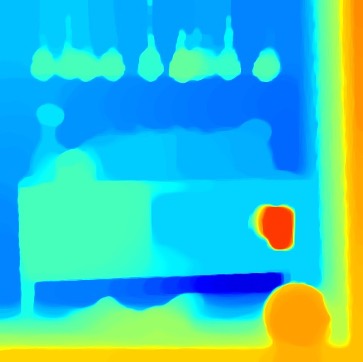} &
    \includegraphics[width=\imagewidth]{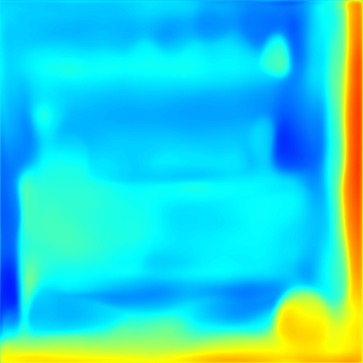} &
    \includegraphics[width=\imagewidth]{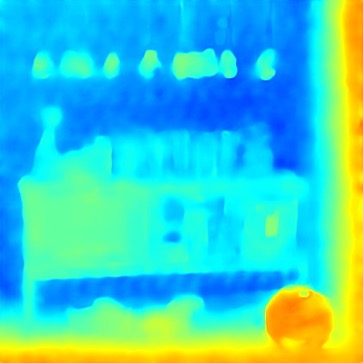} &
    \includegraphics[width=\imagewidth]{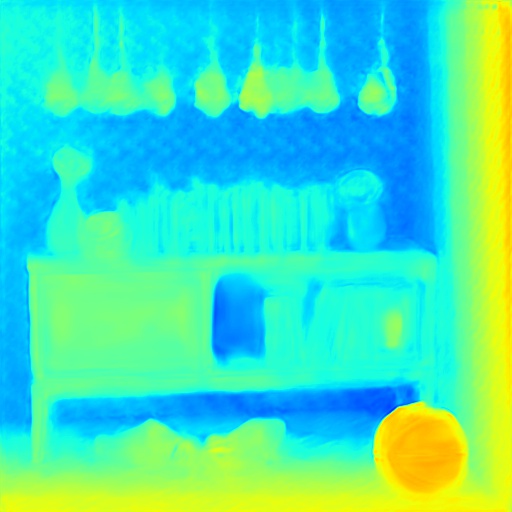} &
    \includegraphics[width=\imagewidth]{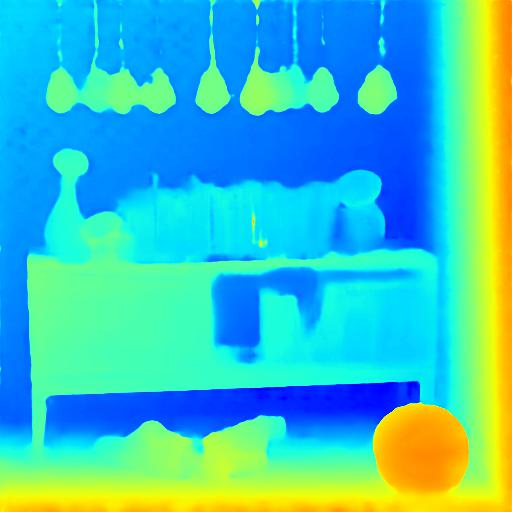} &
    \includegraphics[width=\imagewidth]{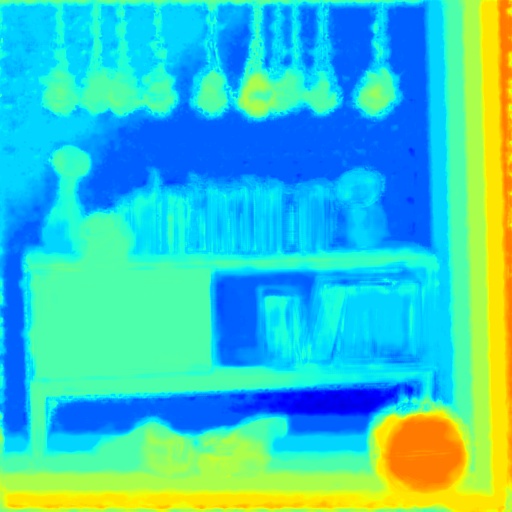} & 
    \includegraphics[width=\imagewidth]{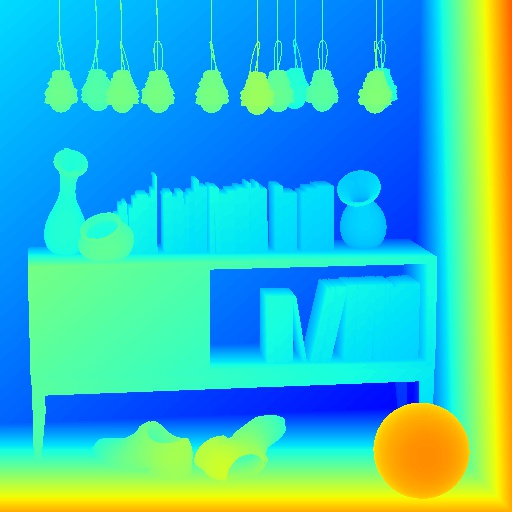}
    \\
    RGB & VDFF~\cite{VDFF} & PSPNet~\cite{PSPNet} & DDFF~\cite{DDFF} & \ren{DefocusNet}~\cite{DefocusNet} & Ours (S) & Ours (US) & GT \\
    \end{tabular}%
    \caption{\textbf{Visual comparison on \ren{4D Light Field Dataset}.} S: \ren{supervised}. US: \ren{ unsupervised (AiF-supervision)}.}
    \label{fig:HCI_baseline}
\end{figure*}
\heading{\ren{4D Light Field Dataset}.}
Compared to \ren{DDFF 12-Scene~\cite{DDFF}, \ren{4D Light Field Dataset}~\cite{honauer2016dataset}} provides the ability to simulate shallower DoF images as its larger baseline leads to a larger synthetic aperture.
Table~\ref{table:HCI_baseline} shows the quantitative \ren{comparison} with state-of-the-art methods.
All of them are trained with this dataset.
Our supervised model achieves the best MSE \ren{(0.0472), which outperforms the state-of-the-art method \ren{DefocusNet}~\cite{DefocusNet} (0.0593)}.
Furthermore, our unsupervised model even performs better than \ren{most of the} other supervised methods, except for \ren{DefocusNet}.

Fig.~\ref{fig:HCI_baseline} shows \ren{the} qualitative results of \ren{4D Light Field Dataset}~\cite{honauer2016dataset}. Our supervised model delivers sharper depth boundaries and less noise in textureless regions. Meanwhile, our unsupervised model also achieves visually comparable results with the supervised DefocusNet~\cite{DefocusNet}.

\begin{table}[]
    \center
    \small
    \caption{
    \textbf{Quantitative comparison on \ren{4D Light Field Dataset}.} \ren{Our supervised model outperforms the state-of-the-art method DefocusNet. Furthermore, our unsupervised model (with AiF supervision) even performs better than most of the other supervised methods.}
    \ren{Please see} Fig.~\ref{fig:HCI_baseline} for the visual comparison. \ren{($*$ represents that the model is pre-trained on \ren{DDFF 12-Scene}.)}
        }
    \renewcommand{\tabcolsep}{4pt} %
    \begin{tabular}{lc|ccc}
        \toprule
        Method & Supervised & MSE $\downarrow$ & RMSE $\downarrow$ & Bump. $\downarrow$\\ 
        \midrule
        Ours & Yes & \best{0.0472} & \best{0.2014} & \second{1.58} \\ 
        \ren{DefocusNet}~\cite{DefocusNet} & Yes & \second{0.0593} & \second{0.2355} & 2.69\\
        *DDFF~\cite{DDFF} & Yes & 0.19 & 0.42 & 1.92\\
        *PSPNet~\cite{PSPNet} & Yes & 0.37 & 0.53 & \best{1.21}\\
        VDFF~\cite{VDFF} & Yes & 1.3 & 1.15 & \second{1.58}\\
        \midrule
        Ours & No & 0.0746 & 0.2398 & 2.58\\ 
        \bottomrule

    \end{tabular}
    \label{table:HCI_baseline}
\end{table}

\heading{\ren{DefocusNet}.}
The \ren{DefocusNet} dataset~\cite{DefocusNet} is a synthetic dataset \ren{using} physically based rendering (PBR) shaders.
Since only a subset of this dataset has been released and \ren{there are no AiF images}, we can only conduct supervised \ren{learning} on the provided subset.
The quantitative results of our method and \ren{DefocusNet}~\cite{DefocusNet} are shown in Table~\ref{table:DefocusNet_baseline}.
Note that the results of \ren{DefocusNet} are retrained on the provided subset with the released code and setting.
\begin{table}[]
    \centering
    \caption{
            \textbf{Quantitative comparison on the \ren{DefocusNet} dataset.} %
            Our supervised model outperforms the state-of-the-art method \ren{DefocusNet}. Please refer to the supplementary material for the full metric results.
            }
    \begin{tabular}{l|ccc}
        \toprule
        {Method} & {MAE}$\downarrow$ & {MSE}$\downarrow$ & {RMSE}$\downarrow$\\ 
        \midrule
        Ours & \textbf{0.0549} & \textbf{0.0127} & \textbf{0.1043}\\ 
        \ren{DefocusNet}~\cite{DefocusNet} &0.0637 & 0.0175 & 0.1207 \\ 
        \bottomrule
    \end{tabular}
    \label{table:DefocusNet_baseline}
\end{table}

\begin{table*}[t!]
    \centering
    \small
    \caption{
            \ren{\textbf{Analysis on generalization ability across different datasets.} We train our models and DefocusNet~\cite{DefocusNet} on FlyingThings3D~\cite{FlyingThings3D} with synthesized focal stacks by the rendering technique used in Barron~\etal~\cite{DBLP:conf/cvpr/BarronASH15}, and test these models on Middlebury Stereo Datasets~\cite{Middlebury} and the DefocusNet dataset. The results show that both of our supervised and unsupervised (with AiF supervision) models have better generalization ability than DefocusNet.}
        }
    \resizebox{\textwidth}{!}{%
    \begin{tabular}{c|c|c|cccccc}
        \toprule
        Method & %
        Supervised 
        & \ren{Test} Dataset &
        \multicolumn{1}{c}{MAE$\downarrow$} & 
        \multicolumn{1}{c}{MSE$\downarrow$} & 
        \multicolumn{1}{c}{RMSE$\downarrow$} & 
        \multicolumn{1}{c}{absRel$\downarrow$} & 
        \multicolumn{1}{c}{sqrRel$\downarrow$} & 
        \multicolumn{1}{c}{Sec.$\downarrow$} 
        \\ 
        \midrule
        Ours & %
        Yes
        & Middlebury &
        \best{3.8249} &
        \best{58.5698} &
        \best{5.9355} &
        \best{2.4776} &
        \best{45.5886} &
        \second{0.0287}
        \\
        Ours &
        No
        & Middlebury &
        \second{5.4499} &
        \second{99.6029} &
        \second{8.2606} &
        3.8954 &
        80.7120 &
        \best{0.0283}
        \\
        \ren{DefocusNet}~\cite{DefocusNet} &%
        Yes
        & Middlebury &
        7.4084 &
        157.4397 &
        9.0794 &
        \second{3.4698} &
        \second{63.6797} &
        0.3798
        \\
        \midrule
        Ours &%
        Yes 
        & \ren{DefocusNet} &
        \second{0.1827} & \second{0.079}5 & \second{0.2607} & \second{72.4664} & \second{40.4281} & \second{0.021}
        \\
        Ours &%
        No 
        & \ren{DefocusNet} &
        \best{0.1816} & \best{0.0627} & \best{0.2380} & \best{59.4459} & \best{14.0227} & \best{0.01959}
        \\
        \ren{DefocusNet}~\cite{DefocusNet} &%
        Yes 
        & \ren{DefocusNet} &
        0.3200 & 
        0.1478 &
        0.3722 &
        138.2917 &
        70.0229 &
        0.0527
        \\
        \bottomrule
    \end{tabular}
    }
    \label{table:middlebury}
\end{table*}

\heading{\ren{Mobile Depth}.}
\ren{The Mobile Depth} dataset~\cite{MobilePhone} is a real-world dataset \ren{of focal stacks captured with mobile phone and cameras}.
Due to the lack of training data in \ren{this} dataset,
all \ren{compared} models are trained on FlyingThings3D~\cite{FlyingThings3D}, \ren{and the input focal stacks are synthesized by the rendering technique used in Barron~\etal~\cite{DBLP:conf/cvpr/BarronASH15}}.

\begin{figure}[]
    \centering
    \footnotesize
    \renewcommand{\tabcolsep}{1pt} %
    \renewcommand{\arraystretch}{1} %
    \newcommand{\imagewidth}{0.33\columnwidth}
    \newcommand{\imageheight}{0.24\columnwidth}
    \resizebox{\linewidth}{!}{%
    \begin{tabular}{cccc}
    \rotatebox[origin=c]{270}{\includegraphics[width=\imagewidth,height=\imageheight]{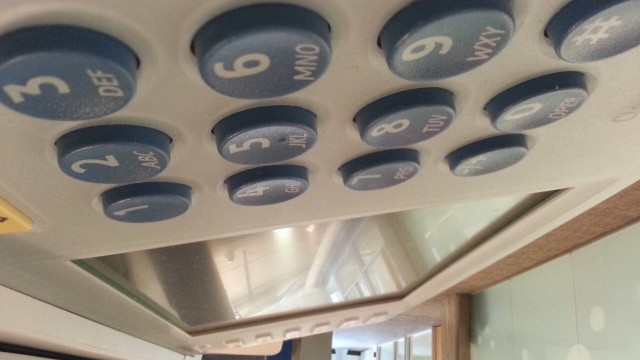}} & 
    \rotatebox[origin=c]{270}{\includegraphics[width=\imagewidth,height=\imageheight]{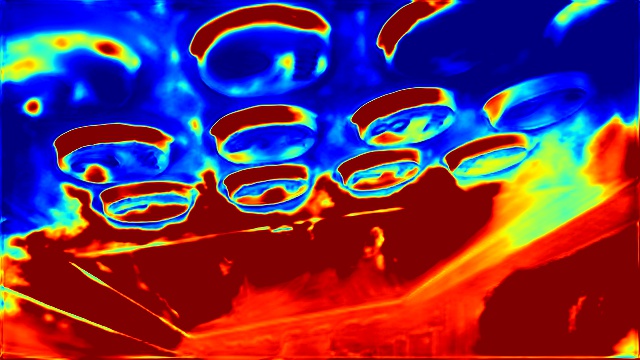}} &
    \rotatebox[origin=c]{270}{\includegraphics[width=\imagewidth,height=\imageheight]{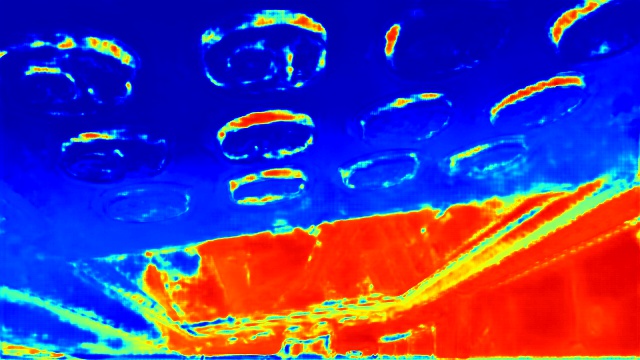}} & 
    \rotatebox[origin=c]{270}{\includegraphics[width=\imagewidth,height=\imageheight]{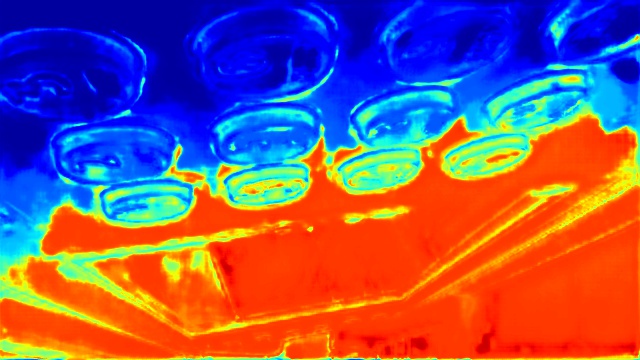}} \\
    AiF GT & \ren{DefocusNet}~\cite{DefocusNet} & Ours (S) & Ours (US) \\
     & \rotatebox[origin=c]{270}{\includegraphics[width=\imagewidth,height=\imageheight]{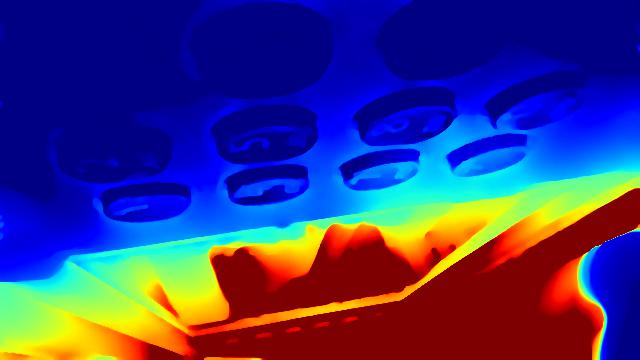}} &
     \rotatebox[origin=c]{270}{\includegraphics[width=\imagewidth,height=\imageheight]{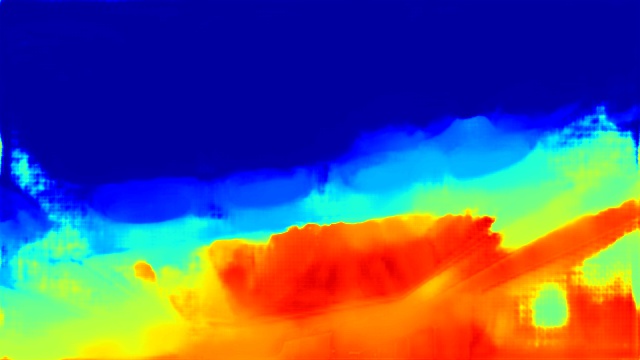}} &
     \rotatebox[origin=c]{270}{\includegraphics[width=\imagewidth,height=\imageheight]{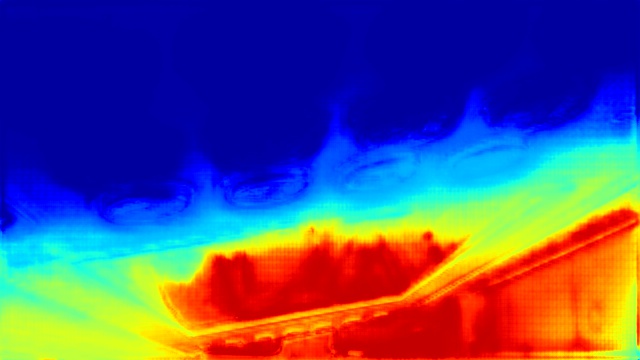}} \\
     & Mobile~\cite{MobilePhone} & Ours (S) * & Ours (US) * \\
    \rotatebox[origin=c]{270}{\includegraphics[width=\imagewidth,height=\imageheight]{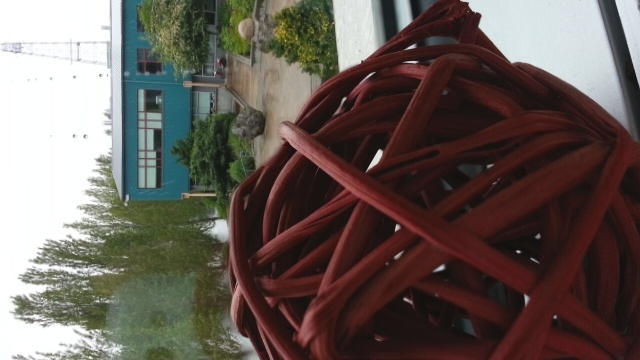}} & 
    \rotatebox[origin=c]{270}{\includegraphics[width=\imagewidth,height=\imageheight]{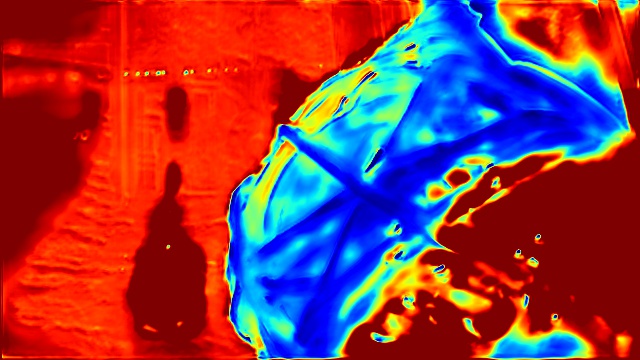}} &
    \rotatebox[origin=c]{270}{\includegraphics[width=\imagewidth,height=\imageheight]{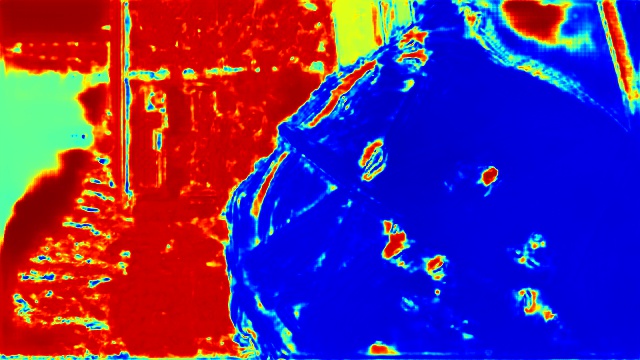}} & 
    \rotatebox[origin=c]{270}{\includegraphics[width=\imagewidth,height=\imageheight]{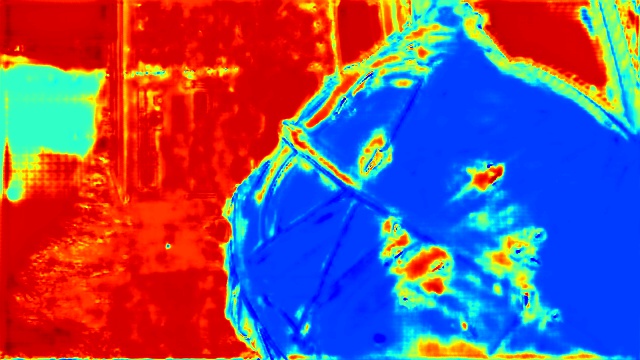}} \\
    AiF GT & \ren{DefocusNet}~\cite{DefocusNet} & Ours (S) & Ours (US) \\
     & \rotatebox[origin=c]{270}{\includegraphics[width=\imagewidth,height=\imageheight]{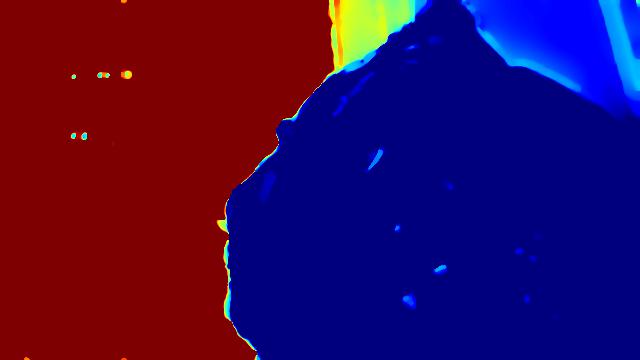}} &
     \rotatebox[origin=c]{270}{\includegraphics[width=\imagewidth,height=\imageheight]{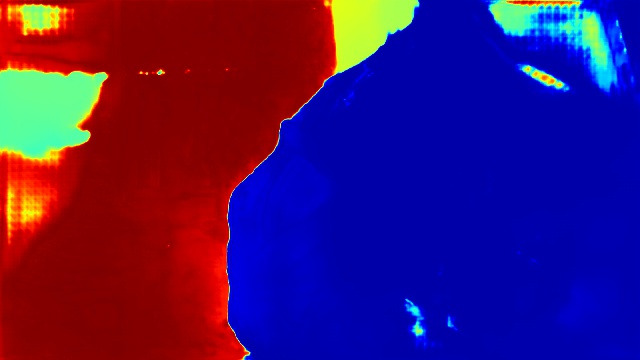}} &
     \rotatebox[origin=c]{270}{\includegraphics[width=\imagewidth,height=\imageheight]{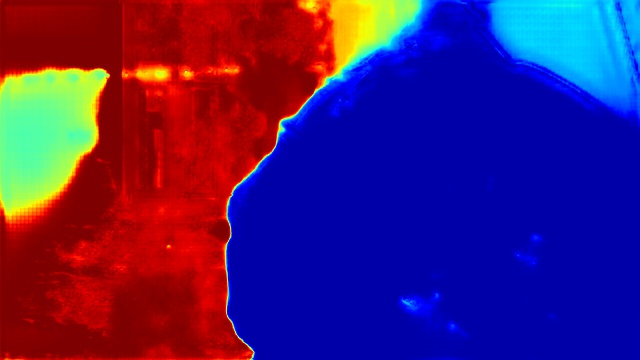}} \\
     & Mobile~\cite{MobilePhone} & Ours (S) * & Ours (US) * \\
    \end{tabular}%
    }
    \caption{\textbf{Visual comparison on the Mobile \ren{Depth} dataset.}
    \ren{With test-time optimization, our models perform better than DefocusNet~\cite{DefocusNet} and favorably against Mobile Depth~\cite{MobilePhone} qualitatively. (S: supervised. US: unsupervised (AiF supervision). *: test-time optimization.)}}
    \label{fig:Mobile_Phone_Fig}
\end{figure}

As shown in Fig.~\ref{fig:Mobile_Phone_Fig}, the output depth maps \ren{of our supervised model}
are smoother and exhibit less ambiguity \ren{than DefocusNet~\cite{DefocusNet}. But the results} of our unsupervised model \ren{(with AiF supervision)} are worse than \ren{DefocusNet}.
\ren{Nonetheless,} since our method allows \albert{training \ren{without ground truth depth},} we can perform \ren{test-time optimization} on this dataset \ren{with} only AiF images.
After \ren{test-time optimization}, our unsupervised model performs better than \ren{DefocusNet, and both of our supervised and unsupervised models perform favorably against Mobile Depth}.
\ren{We also show the output AiF images} in the supplementary material.

\heading{\ren{Generalization ability analysis.}}
\ren{Middlebury Stereo Datasets\cite{Middlebury} is a real-world dataset of stereo images along with ground truth disparity maps. We use this dataset to analyze the generalization ability of our method. To this end, we firstly let our models and DefocusNet~\cite{DefocusNet} be trained on FlyingThings3D~\cite{FlyingThings3D} with synthesized focal stacks by the rendering technique used in Barron \etal \cite{DBLP:conf/cvpr/BarronASH15}, and then test these models on Middlebury Stereo Datasets as well as the DefocusNet dataset. The quantitative results are shown in Table~\ref{table:middlebury}. One can see that both of our supervised and unsupervised (with AiF supervision) models achieve better generalization than DefocusNet on these two datasets.}

\heading{\ren{Running time}.}
As indicated in \ren{Table~\ref{table:middlebury}}, our method is faster than \ren{the state-of-the-art method} DefocusNet~\cite{DefocusNet}. The main reason might be that our model does not have a bottleneck like the global pooling layer in DefocusNet.

\section{Conclusion}

\albert{We \ren{have proposed} a method to jointly estimate the depth map and the all-in-focus (AiF) image from an input focal stack with a shared network. }
By the design of the proposed attention mechanism, the shared network can be trained either supervisedly\ren{, or unsupervisedly with AiF images.}
Our method can \ren{further} mitigate domain gaps \ren{even in the absence of ground truth depth}.
Experimental results show that our method outperforms the state-of-the-art methods while \ren{also having higher efficiency in inference time}.

{\small
\bibliographystyle{ieee_fullname}
\bibliography{egbib}
}

\end{document}